%% file: Kinetal.tex
\documentclass[11pt,a4paper]{article}
\usepackage[hyperref]{acl2021}
\usepackage{times}
\usepackage{latexsym}

\usepackage{booktabs}
\usepackage{float}
\usepackage{graphicx}
\usepackage{amsmath}
\usepackage{amssymb}
\usepackage{amsthm}
\usepackage{amsfonts}
\usepackage{amscd}
\usepackage{amsthm}
\usepackage{amscd}
\usepackage{fontawesome5}
\usepackage{multirow}
\usepackage{caption}
\usepackage{listings}
\usepackage{xcolor}
\usepackage{proof}
\usepackage{pgfplots}
\usepackage{pgfplotstable}
\pgfplotsset{compat=1.17} 
\usepackage{tikz}
\usetikzlibrary{shapes,positioning,calc,backgrounds}
\definecolor{ElectricBlue}{HTML}{2978B5}   
\definecolor{TealGreen}{HTML}{20A39E}     
\definecolor{CoralRed}{HTML}{DF5C61}   
\definecolor{MustardYellow}{HTML}{FFC857}   
\definecolor{VibrantPurple}{HTML}{9B26B6}   
\definecolor{DeepSeaBlue}{HTML}{264653}   
\definecolor{Mint}{HTML}{3EB489}          
\definecolor{LavenderGrey}{HTML}{C4C3D0}  

\lstdefinestyle{prompt}{
    basicstyle=\ttfamily\small,
    breaklines=true,
    frame=single,
    showstringspaces=false
}
\usepackage{microtype}

\aclfinalcopy 

\title{DisCoCLIP: A Distributional Compositional Tensor Network Encoder for Vision-Language Understanding}

\author{Kin Ian Lo \and Hala Hawashin \and Mina Abbaszadeh \\ \textbf{Tilen Limback-Stokin \and Hadi Wazni \and Mehrnoosh Sadrzadeh} \\ University College London}

\date{}

\begin{document}
\maketitle
\begin{abstract}
  Recent vision-language models excel at large-scale image-text alignment but often neglect the compositional structure of language, leading to failures on tasks that hinge on word order and predicate-argument structure. We introduce DisCoCLIP, a multimodal encoder that combines a frozen CLIP vision transformer with a novel tensor network text encoder that explicitly encodes syntactic structure. Sentences are parsed with a Combinatory Categorial Grammar parser to yield distributional word tensors whose contractions mirror the sentence's grammatical derivation. To keep the model efficient, high-order tensors are factorized with tensor decompositions, reducing parameter count from tens of millions to under one million. Trained end-to-end with a self-supervised contrastive loss, DisCoCLIP markedly improves sensitivity to verb semantics and word order: it raises CLIP's SVO-Probes verb accuracy from 77.6\% to 82.4\%, boosts ARO attribution and relation scores by over 9\% and 4\%, and achieves 93.7\% on a newly introduced SVO-Swap benchmark. These results demonstrate that embedding explicit linguistic structure via tensor networks yields interpretable, parameter-efficient representations that substantially improve compositional reasoning in vision-language tasks.

\end{abstract}
\section{Introduction}

Vision-language understanding is a key challenge in AI, with applications to image captioning and multimodal retrieval. Models like OpenAI's CLIP~\cite{radford2021learning} have shown that large-scale joint embeddings can effectively connect visual and textual data.
However, these models mainly rely on Transformer architectures with dense attention, which may overlook the linguistic structure.
%
%
%
%
For instance, recent evaluations of CLIP-like models show that they often ignore word order, acting like bags-of-words  \cite{ThrushJBSWKR22,jiang-etal-2024-comclip,li-etal-2024-interpretable}. The Attribution, Relation and Order (ARO) benchmark~\cite{yuksekgonul2023when} checks if they are able to understand the correct word order. Similarly, the SVO-probes benchmark~\cite{hendricks2021probing} tests if these models mainly focus on nouns, or are also able to recognise verbs. Both of these issues have been common challenges for vision-language models. 

It has been argued that these challenges stem from CLIP-like models being trained on web-sourced image-caption pairs, where captions (often alt-texts) frequently ignore word order and verb usage.
As a result, their contrastive learning is not sensitive to linguistic structure \cite{yuksekgonul2023when}. While training with \emph{hard negatives} could address this, such samples are costly to source. 
Instead, we introduce \textbf{DisCoCLIP}, the first model for vision and language with a text encoder that fully incorporates the compositional linguistic structure of text with the distributions of the words therein.
To achieve this, we represent sentences as tensor networks, where each word is encoded as a tensor and  interactions between words are captured through a series of tensor contractions.

The advantages of using a tensor network text encoder are twofold. First, it enables explicit encoding of both syntactic structure and statistical semantic information, making the resulting text representations more interpretable than those produced by transformer-based encoders. Second, tensor network decompositions can dramatically reduce the number of parameters required, allowing for efficient modelling of high-order interactions without incurring exponential growth in tensor size.  Tensor networks are widely used in  quantum machine learning to capture higher order data correlations~\cite{biamonte2017quantum,schuld2015introduction,stoudenmire2016supervised,cichocki2016tensor}.  Their use in vision-language tasks  might lead to further   advantages coming from the quantum world.

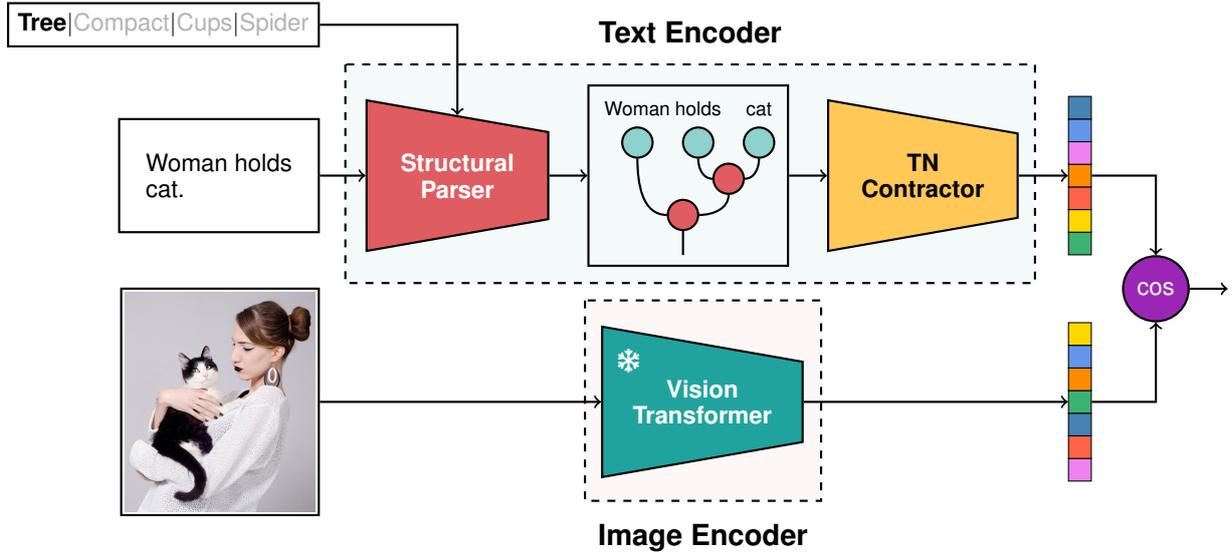
\begin{figure*}[h]
  \centering
  \input{figures/discoclip_arch.tex}
   \caption{\small An illustration of the architecture of DisCoCLIP, which consists of a text encoder based on a structure-informed tensor network of words, and a vision encoder based on a Vison Transformer (ViT).
   The Structural Parser converts the input text into a tensor network, based on the chosen structure which could be any of the four types: \textbf{Tree}, \textbf{Compact}, \textbf{Cups} or \textbf{Spider}. The tensor network is then contracted by the Tensor Network Contractor, which computes an optimal contracting order to obtain a single vector representing the meaning of the input text.
   The input image is processed by Vision Transformer (ViT) to obtain a vector representation of the image.
   The text and image vectors are then used to compute a similarity score, which is used for training the model and for downstream evaluation.
   }
    \label{fig:clip}
\end{figure*}

\textbf{DisCoCLIP} was evaluated on two existing benchmarks on compositional capability: SVO-Probes and  ARO, as well as on  a new SVO-Swap benchmark created by swapping subjects and objects. We compare the performance of \textbf{DisCoCLIP} with CLIP, OpenCLIP \cite{ilharco2021} and BLIP \cite{li2022blip} on these benchmarks.

DisCoCLIP outperforms CLIP and OpenCLIP on verb understanding by $4.82\%$  and $1.01\%$. It also outperforms CLIP in overall performance by $1.3\%$, but falls behind OpenCLIP and BLIP by $2.05\%$ and $7.9\%$. On SVO-Swap, it achieves an accuracy of $93.68\%$ outperforming all three of CLIP, OpenCLIP and BLIP by a large margin ($30.52\%-57.04\%$). On ARO-Relation, again it outperforms all three of the CLIP models by  $4.28\%-5.1\%$, in ARO-Attribution, it outperforms CLIP and OpenCLIP by $9.01\%$ and $10.88\%$, but falls behind BLIP by $8.45\%$. 

In summary, DisCoCLIP achieves comparable performance to transformer-based models with orders of magnitude fewer parameters. The use of tensor decomposition enables efficient representation and computation, making our model more parameter-efficient and potentially more robust when training data is limited.
To our knowledge, it is the first time that  the theory of tensor networks has been used to model the structure of language or used in vision-language tasks. Our work provides a new witness  for  the applications of tensor networks to machine learning and further showcases the advantage of using them.  

\section{Related Work}

Several approaches have been proposed to address these challenges in vision-language models. Some incorporate aspects of linguistic structure~\cite{jiang-etal-2024-comclip}, others introduce hard negatives~\cite{li-etal-2024-interpretable}, and some incentivize learning by explicitly rewarding the model for capturing linguistic elements such as adjectives and verbs~\cite{ThrushJBSWKR22}.

Tensor networks were introduced to make the numerical treatment of many-body quantum states feasible by exploiting their internal structure \cite{white1992}. 
Such states naturally live in exponentially large tensor‐product spaces, which are difficult to handle directly.
A tensor network circumvents this by factorizing a single, high‐order tensor into a set of lower‐order tensors, whose indices are glued together by contraction operations.
In a \emph{Tensor Train} (also known as a Matrix Product State, or MPS), these tensors are arranged in a strictly one-dimensional sequence, with each tensor contracted only to its immediate predecessor and successor through shared \emph{bond} indices; by contrast, a Tree Tensor Network connects tensors in a branching, hierarchical structure.
Tensor networks have found applications outside physics, especially in machine learning where they are used for sequence modelling \cite{Harvey_Yeung_Meichanetzidis_2025}, optimizing the computations of neural networks \cite{Jahromi_Orus_2024,Novikov2015}, and in general any large-scale optimization problem \cite{Cichocki2017}, such as latent feature extraction \cite{Stoudenmire2018} and security \cite{AIZPURUA2025130211}. Their decomposition methods have been tested on image classification tasks \cite{roberts2019,rao2020,serafini2017}, word statistics, and document retrieval from large corpora of text \cite{miller2021tensor,zhang2019,LiuZYCLBC05,bouchard2015}. 

Tensors and the contraction operation between them were also used in a model of meaning known as ``compositional  distributional semantics'' \cite{baroni-zamparelli-2010-nouns,maillard-etal-2014-type,grefenstette-sadrzadeh-2011-experimental,yeung-kartsaklis-2021-ccg}. In this model, the  meaning of each word is either a vector or a higher-order tensor. The orders of the tensors are determined by the grammatical roles of words. Meanings of nouns are vectors, where as meanings of words with functional roles such as adjectives and verbs are matrices and cubes.   
\textbf{DisCoCLIP} is inspired by compositional distributional semantics and the theory of tensor networks. We denote the meaning of a piece of text by a tensor network. In this tensor network,  the tensors encode meanings of words, the layout of the tensor network represents the syntactic structure of the sentence. Other tensor network layouts are used as baselines to test how useful is encoding less structure, such as word order and bags-of-words.  

Another key novelty of our model is that it extends compositional distributional semantics to a multimodal setting. Previous multimodal adaptations include \cite{lewis-etal-2024-clip} for compositional concept learning,  \cite{nazir-sadrzadeh-2024-adjective} for audio-text retrieval, and \cite{wazniverbclip} for verb understanding in CLIP. However, \textbf{DisCoCLIP} differs from these approaches in two important ways. First, our model is more general: It handles sentences of arbitrary syntactic structure, whereas prior work typically focuses on specific constructions such as subject-verb-object \cite{lewis-etal-2024-clip,wazniverbclip} or adjective-noun pairs \cite{nazir-sadrzadeh-2024-adjective}. Second, \textbf{DisCoCLIP} features an end-to-end pipeline trained with a single objective function, in contrast to previous methods that require a separate objective for their different text and audio/image model components. This unified approach enables more flexible and scalable multimodal learning.

\section{Basics of Tensor Networks}

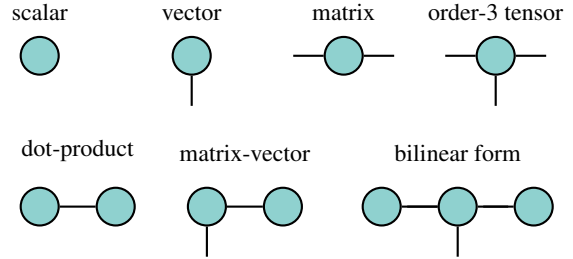
\begin{figure}[h]
\centering
\begin{tikzpicture}[
    tensor/.style={
      draw,
      circle,
      fill=TealGreen!50,
      minimum size=5mm,
      inner sep=0pt,
      thick,
    },
    edge/.style={
      thick,
    },
    every node/.style={font=\small}
  ]
  \node[tensor] (S) {};
  \node[above=2pt of S] {scalar};

  \node[tensor] (V) at (2,0) {};
  \draw[edge] (V.south) -- ++(0,-0.4);
  \node[above=2pt of V] {vector};

  \node[tensor] (M) at (4,0) {};
  \draw[edge] (M.west) -- ++(-0.4,0);
  \draw[edge] (M.east) -- ++(0.4,0);
  \node[above=2pt of M] {matrix};

  \node[tensor] (T) at (6,0) {};
  \draw[edge] (T.west)  -- ++(-0.4,0);
  \draw[edge] (T.south) -- ++(0,-0.4);
  \draw[edge] (T.east)  -- ++(0.4,0);
  \node[above=2pt of T] {order-3 tensor};

  \node[tensor] (DP1)     at (0,-2) {};
  \node[tensor] (DP2)     at (1,-2) {};
  \draw[edge] (DP1.east)  -- (DP2.west);
  \node[above=2pt] at (0.5,-1.6) {dot‐product};

  \node[tensor] (MV1)     at (2.2,-2) {};
  \node[tensor] (MV2)     at (3.2,-2) {};
  \draw[edge] (MV1.east)  -- (MV2.west);
  \draw[edge] (MV1.south) -- ++(0,-0.4);
  \node[above=2pt] at (2.7,-1.6) {matrix‐vector};

  \node[tensor] (MM1)     at (4.5,-2) {};
  \node[tensor] (MM2)     at (5.5,-2) {};
  \node[tensor] (MM3)     at (6.5,-2) {};
  \draw[edge] (MM1.east)  -- (MM2.west);
  \draw[edge] (MM2.west) -- ++(-0.4,0);
  \draw[edge] (MM2.east) -- ++(0.4,0);
  \draw[edge] (MM2.south) -- ++(0,-0.4);
  \draw[edge] (MM3.west) -- ++(-0.4,0);
  \node[above=2pt] at (5.5,-1.6) {bilinear form};
\end{tikzpicture}
\caption{Graphical representation of tensor networks. A tensor is depicted as a node with one edge for each index of the tensor. For example a scalar has no edge, a vector has one edge, a matrix has two edges and an order-3 tensor has 3 edges. An edge of a node can be connected to another edge of another node, forming a \emph{contraction}, which is a generalised form of matrix multiplication.}
\label{fig:tn_samples}
\end{figure}
\noindent
A tensor network is a collection of tensors contracted together to form a new tensor. An order-$n$ tensor $T$ is a multi-dimensional array $T \in \mathbb{R}^{d_1 \times \cdots \times d_n}$, where $d_i$ is the dimension of the $i$-th index. Elements are denoted by $T_{i_1, \ldots, i_n}$, with each $i_k$ ranging from $0$ to $d_{k-1}$. Scalars, vectors, and matrices are tensors of order 0, 1, and 2, respectively.

\paragraph{Tensor contractions.}
Tensors can be multiplied together by contracting over a shared index, which generalizes matrix multiplication. For example, given two tensors $A \in \mathbb{R}^{d_1 \times d_2}$ and $B \in \mathbb{R}^{d_2 \times d_3}$, their contraction over the second index yields a new tensor $C \in \mathbb{R}^{d_1 \times d_3}$:
\[
C_{i_1, j_1} = \sum_{k=1}^{d_2} A_{i_1, k} \; B_{k, j_1}
\]
This operation extends naturally to higher-order tensors by summing over any shared index.

\[
\begin{split}
C_{i_1,\ldots,i_p,j_1,\ldots,j_q}
=\sum_{k_1,\ldots,k_r} &
   A_{i_1,\ldots,i_p,k_1,\ldots,k_r} \\
&\times\,
   B_{k_1,\ldots,k_r,j_1,\ldots,j_q}
\end{split}
\]
where the indices $k_1, \ldots, k_r$ are summed over, representing the contracted dimensions shared by $A$ and $B$. This operation generalizes matrix multiplication and inner product to higher-order tensors.

\paragraph{Graphical representation.}
Tensor contractions involving multiple tensors can be difficult to reason about. The graphical representation of tensors, as shown in Figure~\ref{fig:tn_samples}, provides a more intuitive way of visualizing them. In this representation, tensors are depicted as nodes and their indices as edges, with edges common to two tensors indicating a contraction.

\paragraph{Tensor decomposition.}
As the number of parameters grows exponentially with the tensor order, computing with them becomes costly. Tensor networks were originally introduced to efficiently represent high-order tensors by decomposing them into a network of lower-order tensors. The number of parameters of an order-$n$ tensor $T \in \mathbb{R}^{d_0 \times d_1 \times d_2 \times \cdots \times d_n}$ is given by the product of its dimensions, $d_0 d_1 d_2 \cdots d_n$. 

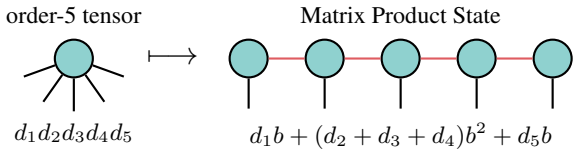
\begin{figure}[h]
\centering
\begin{tikzpicture}[
    tensor/.style={
      draw,
      circle,
      fill=TealGreen!50,
      minimum size=5mm,
      inner sep=0pt,
      thick,
    },
    edge/.style={
      thick,
    },
    every node/.style={font=\small}
  ]

  \node[tensor] (MV1) at (2.2,-0) {};
  \node[above=2pt of MV1] {order-5 tensor};

  \node at (3.5, 0) {\large $\longmapsto$};
  
  \foreach \i in {0,1,2,3,4}{
    \draw[edge] (MV1) -- ++({200 + \i*35}:0.7);
    \node[tensor] (NN\i) at (4.5 + \i, 0) {};
    \draw[edge] (NN\i.south) -- ++ (0, -0.4);
  }
  \foreach \i/\j in {0/1,1/2,2/3,3/4}{
    \draw[edge,CoralRed] (NN\i.east) -- (NN\j.west);
  }
  
  \node[above=2pt of NN2] {Matrix Product State};
  \node[below=13pt of MV1] {$d_1d_2d_3d_4d_5$};
  \node[below=13pt of NN2] {$d_1 b + (d_2+d_3+d_4) b^2 + d_5 b$};
\end{tikzpicture}
\caption{The decomposition of an order-5 tensor into a Matrix Product State (MPS). The red edges are called the bonds and their dimension is called the \emph{bond dimension} $b$.
Below the tensors, we show the formulas for the number of parameters required to represent the full order-5 tensor (bottom left) and its MPS decomposition (bottom right). }
\label{fig:mps}
\end{figure}
In many practical scenarios, representing a high-order tensor with all of its exponentially many parameters is unnecessary. Instead, the tensor can often be efficiently approximated or even exactly represented by decomposing it into a network of lower-order tensors. This decomposition, called a tensor network, greatly reduces the number of parameters and enables scalable computation.

A canonical example is the ground state of a quantum many-body system, which can be efficiently represented by a Matrix Product State (MPS)~\cite{Fannes1992FCS}, also known as a Tensor Train. An MPS expresses a high-order tensor as a sequence (or ``train'') of lower-order tensors connected by contracted indices, as illustrated in Figure~\ref{fig:mps}. 
The contractions between neighboring tensors are called \emph{bonds} and their dimensions are called \emph{bond dimensions}. 
The dimension of each bond index is the \emph{bond dimension} $b$, which controls the expressiveness and parameter count of the MPS. The total number of parameters in the MPS is 
\[
(d_1 + d_n) b + \sum_{k=2}^{n-1} d_k b^2,
\]
assuming all bond dimensions are equal to $b$. This is typically much smaller than the $d_1 d_2 \cdots d_n$ parameters required for a full tensor, making MPS an efficient representation for high-order tensors.
For the rest of this paper, we will use a uniform dimension denoted by $d$ and a uniform bond dimension  denoted by $b$ for simplicity. We denote the number of parameters in an MPS representation of an order-$n$ tensor as 
\begin{equation}
\label{eq:mps}
\#\text{MPS}(n, d, b) = 
\begin{cases}
  d, & n = 1 \\
  2db + (n-2)db^2, & n \geq 2
\end{cases}
\end{equation}

Other than MPS, other common tensor network decompositions include the Tree Tensor Network (TTN)~\cite{Shi2006TTN}, which arranges tensors in a tree structure, and the Projected Entangled Pair State (PEPS)~\cite{Verstraete2004PEPS}, which arranges tensors in a 2D lattice. 
These decompositions are useful for different applications and can be adapted to specific data structures. 

\section{Methodology}
\begin{figure}

\centering
\begin{tikzpicture}[
    tensor/.style={
      draw,
      circle,
      fill=TealGreen!50,
      minimum size=5mm,
      inner sep=0pt,
      thick,
    },
    edge/.style={
      thick,
    },
    every node/.style={font=\small},
    scale=0.9,
    transform shape
  ]
    \node[tensor] (TA) at (0, 0) {};
    \node[tensor] (TL) at (1, 0) {};
    \node[tensor] (TB) at (2, 0) {};
    \node[tensor,fill=CoralRed] (TU1) at (1.5, -0.6) {};
    \node[tensor,fill=CoralRed] (TU2) at (0.75, -1.2) {};
    \node[above=2pt of TA] {Alice};
    \node[above=2pt of TL] {loves};
    \node[above=2pt of TB] {Bob};
    \node at (1, -2.2) {\normalsize\textbf{Tree}};
    \draw[edge] (TU2.south) -- ++(0,-0.4);

    \def\angleA{0}
    \def\angleB{180}
    \draw[edge] (TL.south) to[out=-90, in=\angleB] (TU1.\angleB);
    \draw[edge] (TB.south) to[out=-90, in=\angleA] (TU1.\angleA);
    \draw[edge] (TU1.south) to[out=-90, in=\angleA] (TU2.\angleA);
    \draw[edge] (TA.south) to[out=-90, in=\angleB] (TU2.\angleB);

    \node[tensor] (SA) at (4, 0) {};
    \node[tensor] (SL) at (5, 0) {};
    \node[tensor] (SB) at (6, 0) {};
    \node[tensor,fill=black,minimum size=2.5mm] (SU) at (5, -1) {};
    \node[above=2pt of SA] {Alice};
    \node[above=2pt of SL] {loves};
    \node[above=2pt of SB] {Bob};
    \node at (5, -2.2) {\normalsize\textbf{Spider}};
    \draw[edge] (SU.south) -- ++(0,-0.4);

    \draw[edge] (SA.south) to[out=-90, in=\angleB] (SU.\angleB);
    \draw[edge] (SL.south) to[out=-90, in=90] (SU.90);
    \draw[edge] (SB.south) to[out=-90, in=\angleA] (SU.\angleA);

    \node[tensor] (CA) at (0, -3.5) {};
    \node[tensor] (CL) at (1, -3.5) {};
    \node[tensor] (CB) at (2, -3.5) {};
    \node[above=2pt of CA] {Alice};
    \node[above=2pt of CL] {loves};
    \node[above=2pt of CB] {Bob};
    \node at (1, -4.5) {\normalsize\textbf{Compact}};

    \draw[edge] (CA.-45) to[out=-45, in=-135] (CL.-135);
    \draw[edge] (CB.-135) to[out=-135, in=-45] (CL.-45);
    \draw[edge] (CL.south) -- ++(0,-0.4);

    \node[tensor,fill=MustardYellow] (KS) at (3.5, -3.5) {};
    \node[tensor] (KA) at (4.5, -3.5) {};
    \node[tensor] (KL) at (5.5, -3.5) {};
    \node[tensor] (KB) at (6.5, -3.5) {};
    \node[above=2pt of KS] {\texttt{START}};
    \node[above=2pt of KA] {Alice};
    \node[above=2pt of KL] {loves};
    \node[above=2pt of KB] {Bob};
    \node at (5, -4.5) {\normalsize\textbf{Cups}};

    \draw[edge] (KS.-45) to[out=-45, in=-135] (KA.-135);
    \draw[edge] (KA.-45) to[out=-45, in=-135] (KL.-135);
    \draw[edge] (KL.-45) to[out=-45, in=-135] (KB.-135);
    \draw[edge] (KB.south) -- ++(0,-0.4);

\end{tikzpicture}
   \caption{\small The four types of tensor networks considered in this paper: \textbf{Compact} and \textbf{Tree} are based on the CCG grammar, \textbf{Cups} preserves word order and \textbf{Spider} is a bag-of-words model. Each rectangle represents a node in the tensor network. The black dot in Spider is  the copy node, which is operationally equivalent to element-wise multiplication.
   }
    \label{fig:tensornetworks}
\end{figure}
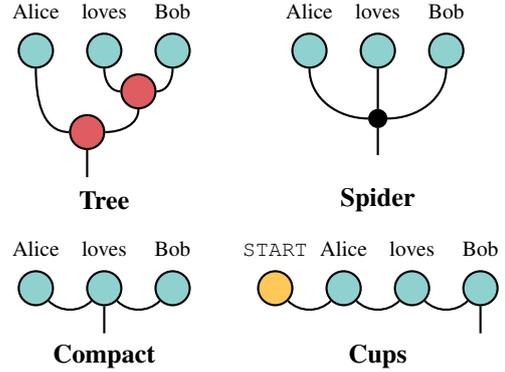

\noindent
Our main contribution is to replace CLIP's Transformer-based text encoder with a  tensor network encoder, resulting in a new vision-language model we call \textbf{DisCoCLIP}. 
In DisCoCLIP, the text encoder constructs sentence embeddings using tensor networks that explicitly encode linguistic structure, while the image encoder remains the original CLIP vision transformer. By varying the layout of the tensor network, we can control the level of syntactic and semantic information captured in the text representation.

Given an image-caption pair, \textbf{DisCoCLIP} processes them in the following steps (see Figure~\ref{fig:clip}):
\hspace{-0.3cm}\begin{enumerate}
    \item The sentence is parsed to extract its syntactic structure. \vspace{-0.3cm}
    \item A tensor network is constructed based on the parse tree, where each word is represented by a tensor node.\vspace{-0.3cm}
    \item The tensor network is contracted to produce a fixed-size vector embedding for the entire sentence.\vspace{-0.3cm}
    \item The image embedding is computed using a Vision Transformer (ViT).\vspace{-0.2cm}
    \item The text and image embeddings are compared to compute a similarity score, which is used for training the model and for downstream evaluation.\vspace{-0.3cm}
\end{enumerate}


For step 1, we use the state-of-the-art \texttt{BobcatParser} \cite{clark2021} from the \texttt{Lambeq} library \cite{kartsaklis2021lambeq} to obtain the Combinatory Categorial Grammar parse trees of the sentences \cite{ades1982,steedman1987,steedman2000} .

Combinatory Categorial Grammar (CCG) is a highly expressive formalism for modeling natural language syntax and semantics. In CCG, each word is assigned a syntactic category that reflects both its grammatical role and its combinatory potential. Categories are either atomic (such as noun phrase $NP$ or sentence $S$) or functional, where functional categories specify how a word combines with its arguments. Functional types take the form $Y/X$ or $Y\backslash X$, indicating that the word expects an argument of type $X$ to its right ($/$) or left ($\backslash$), and yields a result of type $Y$. For example, adjectives have type $NP/NP$, intransitive verbs have type $S\backslash NP$, and transitive verbs have type $(S\backslash NP)/NP$. 

The combinatory rules of CCG allow for the composition of these categories to cancel out the functional types and yield a sentence $S$ type. The two main rules are forward application ($>$) and backward application ($<$):
\[
\infer[>]{X}{X/Y \quad Y}
\qquad
\infer[<]{X}{Y \quad X\backslash Y}
\]
where $X$ and $Y$ are any CCG types. These rules allow for the composition of words into phrases and sentences, following the syntactic structure of the language. For example, the sequence ``Alice loves Bob'' can be reduced to a sentence $S$ by first assigning the atomic category $NP$ to both ``Alice'' and ``Bob'', and the functional category $(S \backslash NP)/NP$ to ``love'' and then applying the forward and backward application rules as follows:  
\[
\infer[<]{S}{\infer{NP}{Alice} & \infer[>]{S\backslash NP}{
  \infer{(S \backslash NP)/NP}{loves} \qquad \infer{NP}{Bob}}
  }
\]
Other CCG rules include forward and backward composition, which are used to combine auxiliary verbs with their arguments, and forward and backward cross-composition, used to combine categories with long distance dependencies such as gapping. Another notable CCG rule is type-raising, which enables specific combinations of categories, e.g. from left to right. This feature helps the CCG align with Psycholinguistic theories. For instance, in English, it will allow categories to combine from left to right and form incremental parses that support theories of human sentence processing. 

A distributional compositional (DisCo) semantics has been developed for CCG \cite{grefenstette-sadrzadeh-2011-experimental,yeung-kartsaklis-2021-ccg,wijnholds-etal-2020-representation}.
This semantics assigns to a word $w$ with a CCG category composed of $n$ atomic categories a multilinear map $f_w$ with $n$ arguments  
\[
f_w \colon V_1 \times V_2 \times \cdots \times V_n \to V_{n+1}
\]
Each $V_i$ is a finite-dimensional vector space over the field of reals $\mathbb{R}$.  Equivalently,  $f_w$ can be represented by a tensor of in the space \[
f_w \in V_1 \otimes V_2 \otimes \cdots \otimes V_{n+1}
\]
Here,  each atomic type corresponds to an index of the tensor. 
For example, a noun with the type $NP$ is assigned a vector (order-1 tensor), while an adjective with the type $NP/NP$ is assigned a linear map that takes a vector and returns a vector, which can be represented as a matrix (order-2 tensor). A transitive verb with the type $(S\backslash NP)/NP$ is assigned a bilinear map that takes two vectors and returns another vector, i.e. a cube (an order-3 tensor), and so on. 
For the general formulae of these representations, see \cite{maillard-etal-2014-type} and \cite{wijnholds-etal-2020-representation}.

Given the CCG parse tree, the word tensors are composed by performing tensor contractions that mirror the syntactic reductions specified by the tree. Each time a combinatory rule (such as forward or backward application) is applied in the parse, the corresponding word tensors are contracted along the appropriate indices. This process recursively combines the tensors according to the grammatical structure, ultimately yielding a single vector representation for the entire sentence.
Such semantics was developed in \cite{maillard-etal-2014-type,wijnholds-etal-2020-representation} and leads to the \textbf{Compact} tensor network structure.

An alternative semantics presented in \cite{yeung-kartsaklis-2021-ccg} assigns  to every word a vector and models the grammatical compositions (represented by CCG rules such as  forward and backward  application) by a shared order-3 tensor. This tensor acts as a universal composition operator of all compositional operators. This approach yields the \textbf{Tree} tensor network structure, where the parse tree topology is preserved but all internal nodes use the same composition tensor to combine their child representations.

\subsection{Text Encoder Structures}
We consider four types of tensor network structures: \textbf{Tree}, \textbf{Compact}, \textbf{Cups}, and \textbf{Spider}, as illustrated in Figure~\ref{fig:tensornetworks}.
Every tensor node in the networks is a trainable parameter, which is learned during the training process.


The \textbf{Tree} structure is based on the CCG parse tree of the sentence, where each word is represented as a vector node and an order-3 tensor is used to compose these word nodes to form non-terminal terms in the parse tree. 

The \textbf{Compact} structure is a variant of the \textbf{Tree} structure, where every non-terminal node in the parse tree is absorbed by one of its parents, resulting in a more compact representation where some word nodes become higher-order tensors.

The \textbf{Cups} structure is a variant of Tensor Train (or MPS) where each word is an order-2 tensor, connected in a chain to preserve word order. The first word connects to a special \texttt{start} node while the last word outputs the sentence embedding.

The \textbf{Spider} structure implements a bag-of-words model, where each word is represented as a vector node and all word nodes are contracted through a special \emph{copy node} to produce a single output vector. This copy node, of order $n$, is a tensor $C \in \mathbb{R}^{d^n}$ defined as
\[
C_{i_1, i_2, \ldots, i_n} = 
\begin{cases}
1 & \text{if } i_1 = i_2 = \cdots = i_n, \\
0 & \text{otherwise.}
\end{cases}
\]
Contracting $n-1$ indices of the copy node with $n-1$ word vectors yields their element-wise (Hadamard) product, producing a multiplicative bag-of-words sentence embedding.

\textbf{Parameter count.} Each tensor network structure has a different parameter count, determined by the number and order of word tensors and any composition tensors. Let $|V|$ be the vocabulary size. For \textbf{Compact}, let $|V^r|$ be the number of words with order-$r$ tensors, and $\#\text{MPS}(r, d, b)$ the parameter count for an order-$r$ MPS (see Eq.~(\ref{eq:mps})). Table~\ref{tab:params} summarizes the counts.

\begin{table}[h]
  \small
  \centering
  \caption{Number of parameters for each tensor network structure.}
  \label{tab:params}
  \begin{tabular}{lccc}
    \toprule
      \textbf{Structure} & \textbf{Words} & \textbf{Composition} \\
      \midrule
      \textbf{Tree}     & $|V|d$         & $2db+db^2$    \\
      \textbf{Compact}  & $\sum_r |V^r|\#\text{MPS}(r, d, b)$         & $0$           \\
      \textbf{Spider}   & $|V|d$         & $0$           \\
      \textbf{Cups}     & $|V|d^2$         & $d$           \\
      \bottomrule
  \end{tabular}
\end{table}

\section{Contrastive Learning}
We train \textbf{DisCoCLIP} using contrastive learning, where the image encoder $f$ (frozen CLIP) and the tensor network text encoder $g$ map image-caption pairs $(x, y)$ to embeddings $(\mathbf{x}, \mathbf{y})$. The goal is to bring true (positive) pairs closer and push (negative) mismatched pairs apart in the joint embedding space.
For a batch of $B$ positive pairs, all non-matching image-caption combinations in the batch ($B(B-1)$) serve as negatives, following the in-batch negative sampling of CLIP~\cite{radford2021learning}. These are considered \emph{easy} negatives, as opposed to more challenging, hand-crafted \emph{hard} negatives.

We use the widely adopted InfoNCE loss~\cite{vandenOord2018} to train the model.
Given a batch of $B$ image-caption pairs with embeddings $\mathbf{x}_i = f(x_i)$ and $\mathbf{y}_i = g(y_i)$, the InfoNCE loss is
\begin{align}
  \mathcal{L} = -\sum_{i=1}^B \log \frac{\exp(s(\mathbf{x}_i, \mathbf{y}_i)/\tau)}{\sum_{j=1}^B \exp(s(\mathbf{x}_i, \mathbf{y}_j)/\tau)},
\end{align}
where $\tau$ is a temperature parameter and $s(\mathbf{x}, \mathbf{y})$ is the cosine similarity between the image embedding $\mathbf{x}$ and the caption embedding $\mathbf{y}$.
Here, the numerator measures similarity for positive (matching) pairs ($i = j$), while the denominator includes all pairs in the batch, serving as negatives when $i \neq j$. The loss thus encourages higher similarity for true pairs and lower for mismatched ones.



\begin{table*}[h]
    \centering
\small
    \caption{Results on the SVO-Probes, SVO-Swap and the ARO datasets. 
    }
  \label{tab:results}
  \begin{tabular}{lcccc|c|cc}
\toprule
& \multicolumn{4}{c|}{\textbf{SVO-Probes}}
& \textbf{SVO-Swap}
& \multicolumn{2}{|c}{\textbf{ARO}} 
\\
& {\scriptsize\textbf{Subject}} 
& {\scriptsize\textbf{Verb}} 
& {\scriptsize\textbf{Object}} 
& {\scriptsize\textbf{Overall}} 
& 
& {\scriptsize\textbf{Attribution}}
& {\scriptsize\textbf{Relation}}
\\
        \midrule
        \textbf{Spider} & 83.29 & 76.48 & 86.64 & 80.95 & 50.00 & 50.00 & 50.00 \\
        \textbf{Cups} & 74.25 & 75.36 & 86.83 & 78.35 & 84.21 & 63.07 & 52.68 \\
        \textbf{Tree} & \underline{89.79} & 79.40 & 85.88 & \underline{83.66} & 47.37 & 55.11 & 52.36 \\
        \textbf{Compact}  & 80.74 & \underline{82.42} & \underline{87.79} & 83.55 & \textbf{93.68} & \underline{70.01} & \textbf{55.81}  \\
        \midrule
        \textbf{CLIP} & 82.83 & 77.60 & 90.08 & 82.36 & 57.89 & 61.00 & 51.53 \\
        \textbf{OpenCLIP} & 85.15 & 81.41 & 93.51 & 85.71 & \underline{63.16} & 59.13 & 50.71  \\
        \midrule
       \textbf{BLIP} & \textbf{91.88} & \textbf{88.58} & \textbf{96.37} & \textbf{91.56} & 36.84 & \textbf{78.46} & \underline{52.90} \\
        \bottomrule
    
    \end{tabular}
    \end{table*}

\noindent
We evaluate our approach on two key benchmarks for vision-language understanding: SVO-Probes and ARO. The SVO-Probes dataset is designed to test whether models can distinguish fine-grained changes in the image which corresponds to variations in subject, verb, or object. The task is to determine which of the two images correctly matches a given caption.
In contrast, the ARO (Attribution, Relation, and Order) dataset assesses a model's ability to correctly compose meanings in a sentence. The task is to determine which of the two captions correctly describes a given image. Together, these datasets provide a comprehensive evaluation of both compositional and structural language understanding in multimodal models.

\begin{figure*}[h!]
\small
\centering
\begin{tabular}{m{3cm}| >{\centering\arraybackslash}m{3cm}| >{\centering\arraybackslash}m{3cm} >{\centering\arraybackslash}m{3cm}}
\textbf{Dataset} & \textbf{Caption} & \faCheck \quad \textbf{Positive image} & \faTimes \quad \textbf{Negative image} \\
\midrule
\textbf{SVO-Probes} & A \textbf{father} holds a baby & \includegraphics[width=3cm]{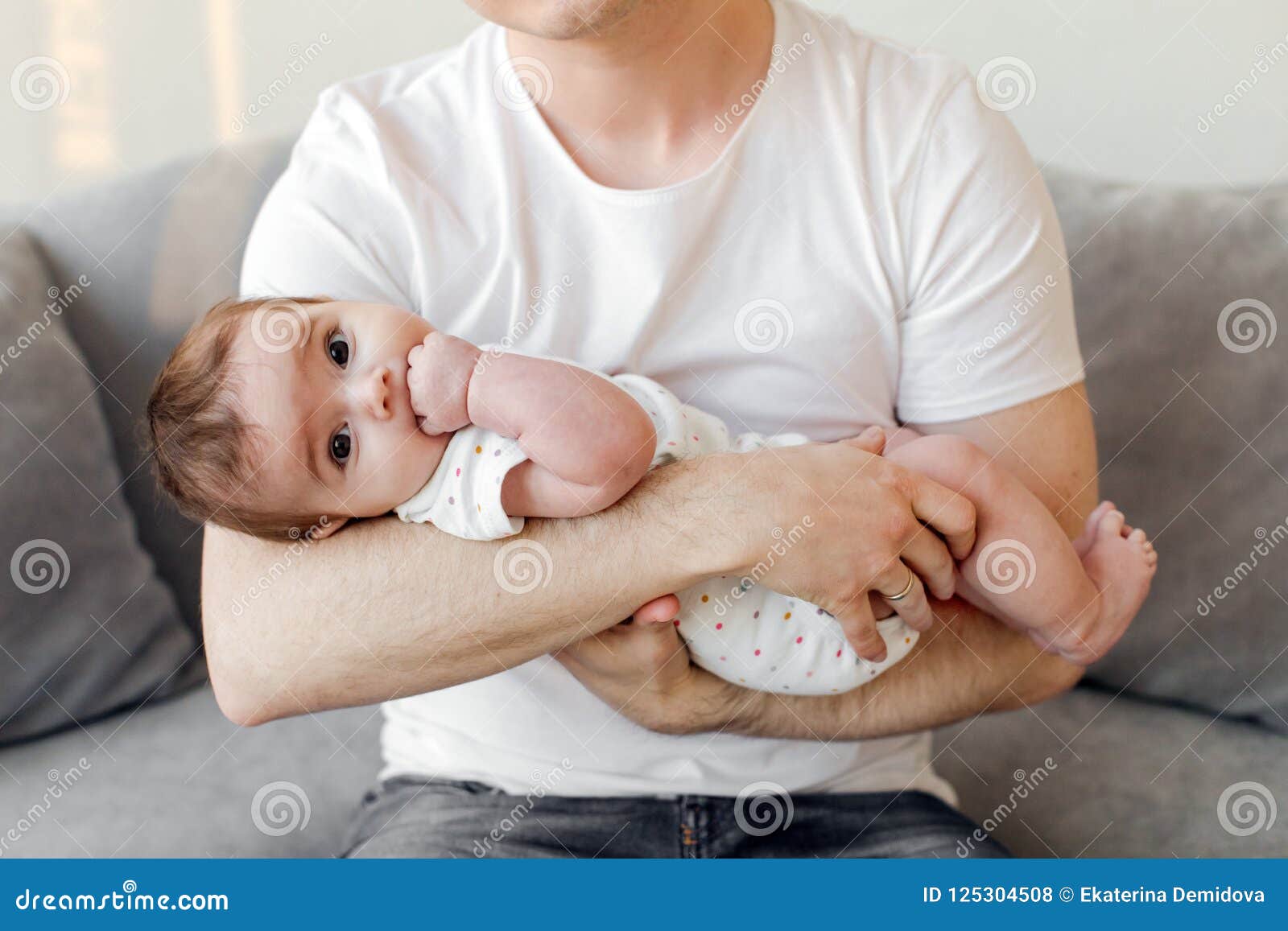} & \includegraphics[width=3cm]{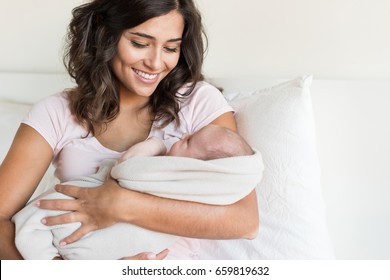} \\
\end{tabular}

\vspace{1.5em}

\begin{tabular}{m{3cm}| >{\centering\arraybackslash}m{3cm}| >{\centering\arraybackslash}m{3cm} >{\centering\arraybackslash}m{3cm}}
\textbf{Dataset} & \textbf{Image} & \faCheck \quad \textbf{Positive caption} & \faTimes \quad \textbf{Negative caption} \\
\midrule
\textbf{SVO-Swap} & \includegraphics[width=3cm, trim=0 8cm 0 1cm, clip]{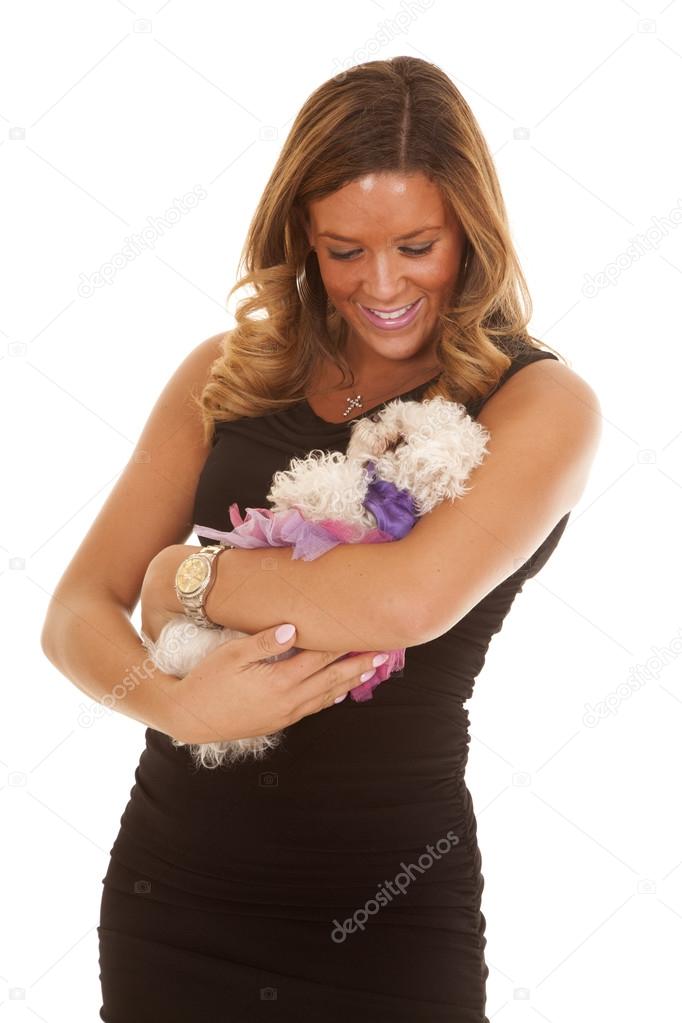}  & A \textbf{woman} holds a \textbf{puppy} & A \textbf{puppy} holds a \textbf{woman} \\
\textbf{ARO-Relation} & \includegraphics[width=3cm]{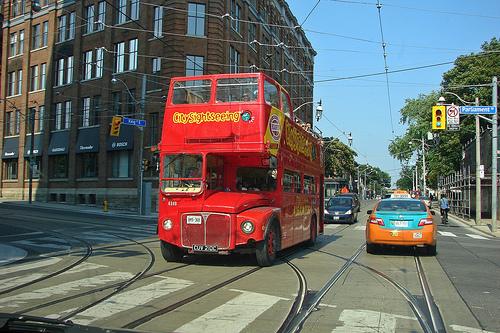}  & The \textbf{bus} is to the right of the \textbf{building} & The \textbf{building} is to the right of the \textbf{bus}\\ 
\textbf{ARO-Attribution} & \includegraphics[width=3cm]{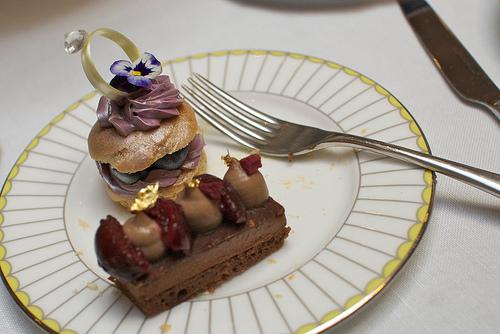}  & The \textbf{dark} brown icing and the \textbf{silver} fork & The \textbf{silver} icing and the \textbf{dark} brown fork \\

\end{tabular}
\caption{Example entries from the datasets used in this work.}
\label{fig:examples}
\end{figure*}
\subsection{SVO-Probes}
For SVO-Probes, we prompted the language model \texttt{Llama-3.2-3B} \cite{llama3} to correct grammatical and spelling mistakes as the original dataset contained errors from crowdsourced captions.
The exact prompt used can be found in the Appendix~\ref{sec:prompt}.
The images in the SVO-Probes dataset were not available for download from the official repository; therefore, we attempted to download them from the Internet using the provided URLs on 2 May 2025. However, many of the URLs were no longer active, and we were only able to download 8,984 images of the total 14,097 images in the dataset, resulting in a reduction of the dataset size from 36,841 to 20,458 entries.

To ensure a reasonable train-test vocabulary overlap, we filtered out entries that contained words that appeared fewer than 50 times in the entire dataset, yielding 8,984 image-caption pairs split 60/20/20 for training, validation, and test, with no image overlap between splits. We also introduced a new dataset: SVO-Swap. This is a set of 95 evaluation pairs created by swapping subjects and objects (when both refer to humans or animals) in SVO-Probes captions. 

The SVO-Probes benchmark is divided into three subsets: Subject, Verb, and Object. Each of these corresponds to the specific component of the sentence that differs between the two alternatives. This structure enables a fine-grained evaluation of the model's ability to distinguish changes in the linguistic roles of the words within a caption.
\subsection{ARO}
The ARO dataset consists of four different subsets: Visual Genome Attribution (VG-A), Visual Genome Relation (VG-R), COCO Order and Flickr Order. 
The way these subsets are constructed was to first gather a set of positive image-caption pairs, and then apply a certain modification to the captions to form negative captions.
In the VG-A subset, positive pairs are chosen to be images with two objects and each gets an attribute. For example \emph{the silver fork and the round plate} contains a fork that is silver, and a plate that is round. The corresponding negative caption would be \emph{the round fork and the silver plate}, where the attributes for the two objects are swapped.
For the VG-R subset, positive pairs are images with a relation involving two objects. For example \emph{}
For ARO-Attribution (with 28,748 entries) and ARO-Relation (with 23,937 entries), we used a 70/15/15 split without frequency filtering, as vocabulary overlap was sufficient.

\subsection{Training}
We trained the model for 10 epochs, using the \texttt{AdamW} optimizer \cite{loshchilov2019} with a learning rate of $10^{-3}$, a weight decay of $10^{-2}$ and a batch size of 64. We also experimented with bond dimensions 2, 5, 10, 15 and 20 in the MPS decomposition.
Training was performed on an Apple M1 MacBook with 16GB RAM, utilizing the PyTorch Metal Performance Shaders (\texttt{mps}) backend to accelerate tensor operations on the GPU.
Each epoch required several minutes, and the total training time for all experiments was approximately one day.
The code used for the experiments is available at \href{https://github.com/kinianlo/discoclip}{github.com/kinianlo/discoclip}.

\subsection{Results}

Table \ref{tab:results} reports our performance on SVO-Probes and ARO. Although BLIP achieves the highest raw scores on SVO-Probes subsets( Subjects (91.88), Verbs (88.58), Objects (96.37)), our \textbf{Compact} model remains a strong second overall (83.55) and is the clear leader among non-BLIP approaches. Notably, \textbf{Compact} scores higher on Verbs (82.42) than on Subjects (80.74), reversing the typical trend seen in all other models and underscoring its structure-aware design for modeling action semantics. On the SVO-Swap benchmark, \textbf{Compact} excels with 93.68, highlighting its robustness to argument perturbations. Finally, on ARO, \textbf{Compact} outperforms every model on Relation attribution (55.81) and closely matches BLIP on Attribution (70.01), demonstrating that embedding syntactic structure as an inductive bias without hard-negative training yields consistently strong relational reasoning and verb understanding.

It is noteworthy that although \textbf{BLIP} achieved the highest overall accuracy on SVO-Probes (91.56), it performed poorly on our new SVO-Swap benchmark (36.84). The underlying causes of this discrepancy remain under investigation.


By contrast, the baseline models \textbf{Spider} and \textbf{Cups} deliver the poorest performance, underscoring that correct structural encoding is essential for compositional understanding. As a bag-of-words model, \textbf{Spider} produces identical representations for both candidate captions in SVO-Swap and ARO, resulting in a flat 50 percent accuracy on these tasks. This failure further illustrates the necessity of incorporating explicit linguistic structure rather than relying solely on word co-occurrence.

\paragraph{Parameter Efficiency}
As shown in Table~\ref{tab:params}, our \textbf{Compact} text encoder requires only 537,600 parameters on the SVO-Probes benchmark—over two orders of magnitude fewer than CLIP's 63,428,097 and BLIP's 137\,258\,496—benefiting from its tensor-train factorization and the relatively small vocabulary size. On the ARO benchmark, \textbf{Compact} uses 28,309,504 parameters—approximately three times fewer than CLIP's text encoder—while still outperforming CLIP in both attribution and relation accuracy.

%
%

\begin{table}[h]
\small
\centering
\caption{Parameter counts for each text encoder model.}
\label{tab:param_counts}
\begin{tabular}{lrr}
\toprule
\textbf{Model} & \textbf{SVO} & \textbf{ARO} \\
\midrule
Spider   & $55,296$ & $735,744$ \\
Cups     & $1,659,392$ & $14,715,392$ \\
Tree     & $185,856$ & $797,184$ \\
Compact  & $537,600$ & $28,309,504$ \\
\midrule
CLIP     & \multicolumn{2}{c}{63,428,097}  \\
OpenCLIP     & \multicolumn{2}{c}{63,428,097} \\
BLIP     & \multicolumn{2}{c}{137,258,496} \\
\bottomrule
\end{tabular}
\end{table}


\begin{figure}
\centering
\begin{tikzpicture}[scale=0.7]
\begin{axis}[
    ybar,
    bar width=8pt,
    width=11cm,
    height=6cm,
    ylabel={Accuracy},
    symbolic x coords={Tree, Compact, Cups, Spider, CLIP, OpenCLIP, BLIP},
    xtick=data,
    ymin=70, ymax=100,
    legend style={at={(0.5,-0.15)}, anchor=north, legend columns=3},
    enlarge x limits=0.1,
    every node near coord/.append style={
    /pgf/number format/fixed,
    /pgf/number format/precision=0
    },
    font=\small,
]
\addplot+[
  ybar, 
  fill=TealGreen, 
  draw=TealGreen,
  nodes near coords,
  nodes near coords style={font=\tiny, text=TealGreen},
  ] coordinates {(Tree, 89.79) (Compact, 80.74) (Cups, 74.25) (Spider, 83.29) (CLIP, 82.83) (OpenCLIP, 85.15) (BLIP, 91.88)};
\addplot+[ybar, fill=CoralRed, draw=CoralRed,
  nodes near coords,
  nodes near coords style={font=\tiny, text=CoralRed},
  ] coordinates {(Tree, 79.40) (Compact, 82.42) (Cups, 75.36) (Spider, 76.48) (CLIP, 77.60) (OpenCLIP, 81.41) (BLIP, 88.58)};
\addplot+[ybar, fill=MustardYellow, draw=MustardYellow,
  nodes near coords,
  nodes near coords style={font=\tiny, text=MustardYellow},
  ] coordinates {(Tree, 85.88) (Compact, 87.79) (Cups, 86.83) (Spider, 86.64) (CLIP, 90.08) (OpenCLIP, 93.51) (BLIP, 96.37)};

\legend{Subject, Verb, Object}
\end{axis}
\end{tikzpicture}
\caption{Performance of models on the SVO-Probes Subject, Verb, and Object subsets.}
\end{figure}
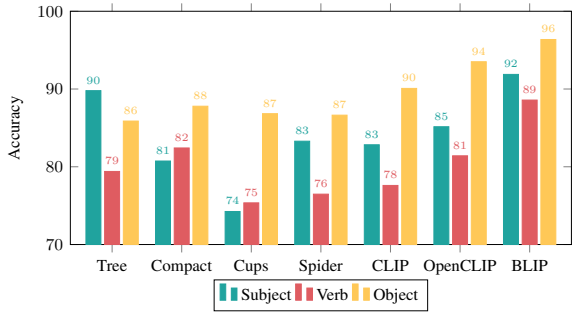

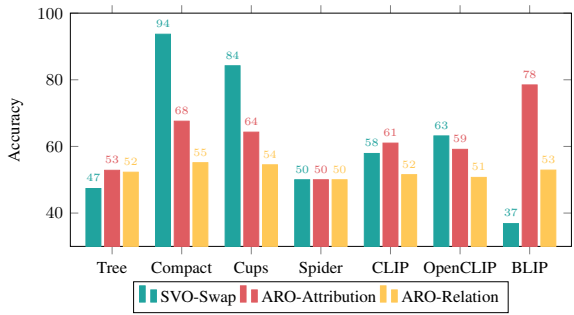
\begin{figure}
\centering
\begin{tikzpicture}[scale=0.7]
\begin{axis}[
    ybar,
    bar width=8pt,
    width=11cm,
    height=6cm,
    ylabel={Accuracy},
    symbolic x coords={Tree, Compact, Cups, Spider, CLIP, OpenCLIP, BLIP},
    xtick=data,
    ymin=30, ymax=100,
    legend style={at={(0.5,-0.15)}, anchor=north, legend columns=3},
    enlarge x limits=0.1,
    every node near coord/.append style={
    /pgf/number format/fixed,
    /pgf/number format/precision=0
    },
    font=\small,
]
\addplot+[
  ybar, 
  fill=TealGreen, 
  draw=TealGreen,
  nodes near coords,
  nodes near coords style={font=\tiny, text=TealGreen},
  ] coordinates {(Tree, 47.37) (Compact, 93.68) (Cups, 84.21) (Spider, 50.00) (CLIP, 57.89) (OpenCLIP, 63.16) (BLIP, 36.84)};
\addplot+[
  ybar, 
  fill=CoralRed, 
  draw=CoralRed,
  nodes near coords,
  nodes near coords style={font=\tiny, text=CoralRed},
  ] coordinates {(Tree, 52.84) (Compact, 67.58) (Cups, 64.29) (Spider, 50.00) (CLIP, 61.00) (OpenCLIP, 59.13) (BLIP, 78.46)};
\addplot+[
  ybar, 
  fill=MustardYellow, 
  draw=MustardYellow,
  nodes near coords,
  nodes near coords style={font=\tiny, text=MustardYellow},
  ] coordinates {(Tree, 52.27) (Compact, 55.12) (Cups, 54.49) (Spider, 50.00) (CLIP, 51.53) (OpenCLIP, 50.71) (BLIP, 52.90)};

\legend{SVO-Swap, ARO-Attribution, ARO-Relation}
\end{axis}
\end{tikzpicture}
\caption{Performance of models on the SVO-Swap and ARO Attribution and Relation benchmarks.}
\end{figure}

\section{Conclusion}
In this work, we introduced {\bf DisCoCLIP}, a vision-language model that replaces the standard Transformer-based text encoder with a structure-informed tensor network. By leveraging the compositional layouts of  tensor networks inspired by compositional distributional semantics and quantum-inspired tensor decompositions, our approach explicitly encodes linguistic structure and achieves competitive performance on challenging multimodal benchmarks such as SVO-Probes and ARO. Our experiments demonstrate that structure-aware tensor networks, particularly the {\bf Compact} model that was a dense variant of the syntactic parse tree, can match or surpass classical neural models in tasks requiring fine-grained understanding of sentence structure. Our model also uses significantly fewer number of parameters in comparison to Transformer-based models such as CLIP. These results highlight the potential of tensor network architectures as interpretable and parameter-efficient alternatives for multimodal learning. Future work will explore scaling these models to larger datasets, working with complex datasets such as Winoground \cite{ThrushJBSWKR22}, exploring the quantum connections and training circuit ansatze, and extending the approach to more complex linguistic phenomena.


\section{Limitations}

A limitation of this work is its evaluation on smaller, curated datasets rather than the web-scale data used to train many contemporary vision-language models. The SVO-Swap benchmark comprises only 95 evaluation pairs. Consequently, the performance reported on this task is not statistically robust.

Our pipeline introduces a dependency on the CCG parser. Errors from the parser can propagate through the pipeline, resulting in ill-formed tensor networks and inaccurate semantic representations.

The image encoder was kept frozen during training, meaning the text encoder learned to align with a fixed set of visual features rather than co-adapting with the image encoder. While this design choice effectively isolates the contribution of the text encoder, 
training both the text and image encoder could potentially yield further performance improvements.

\bibliographystyle{acl_natbib}
\bibliography{ref}

\appendix
\section{Prompt for Grammatical Correction for SVO-Probes}
\label{sec:prompt}
\begin{lstlisting}[style=prompt, caption={Prompt provided to Llama-3.2-3B-Instruct for grammatical correction for caption in the SVO-Probes dataset.}, label={lst:ccg-prompt}]
### System
You are a grammar assistant expert in Combinatory Categorial Grammar.

### Variables
Subject: {subj}
Verb: {verb}
Object: {obj}

### Task
Turn the user's caption fragment into a single English sentence that:
- Is grammatically correct
- Has a valid CCG parse that leads to a sentence output
- Has no spelling errors
- Has a main verb {verb} in simple present tense only
- Has the subject ({subj}) first, followed by the verb ({verb}), then the object ({obj})

Additional rules:
- If the main verb is not {verb}, you may remove parts of the user's input
- If the user's input is a question, convert it into an affirmative sentence.

If it's already correct, repeat it verbatim.
Respond **only** with the final sentence.

### Example
Input: Girl standing in the grass.
Output: The girl stands in the grass.

Input: A person is telling the boy to sit on the chair.
Output: The boy sits on the chair.

Input: The player backhands when he plays tennis.
Output: The player plays a backhand when he plays tennis.

Input: Can we take the kid for a walk on the beach?
Output: The kid walks on the beach.

Input: Is this person resting under the tree?
Output: The person rests under the tree.

### User
{input_sentence}
\end{lstlisting}

\end{document}

%% file: figures/discoclip_arch.tex
\begin{tikzpicture}[
    encoder/.style={draw, shape=trapezium, trapezium angle=80, minimum height=1.5cm, minimum width=2.0cm, align=center, shape border rotate=270, thick},
        input_text/.style={align=center, rectangle, draw, minimum height=1.5cm, thick},
        input_image/.style={draw, thick, inner sep=0.05cm},
        font=\sffamily\small
]
  \node (text_example) [input_text, anchor=east, inner sep=10pt, align=left] at (3, 0) {Woman holds\\cat.};
  \node (image_example) [input_image, anchor=east] at (3, -3) {
\includegraphics[width=2.5cm]{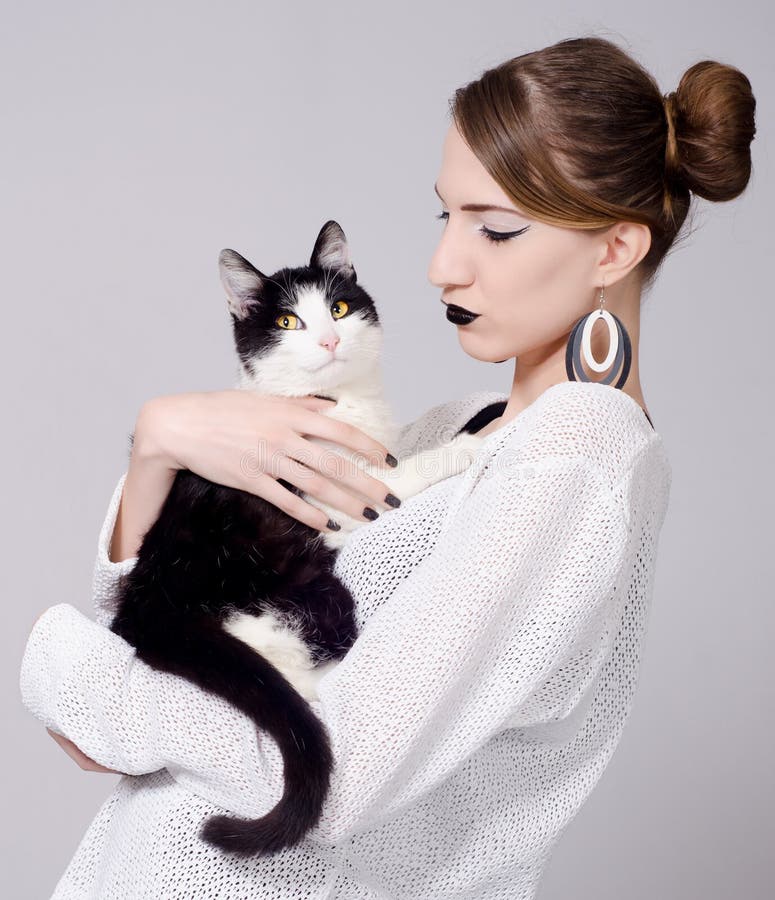}
    };

    \node (text_encoder) [encoder, fill=CoralRed, text=white, anchor=west, right=0.6cm of text_example.east] {\textbf{Structural}\\\textbf{Parser}};
    \node (image_encoder) [encoder, fill=TealGreen, text=white, anchor=west, right=3.7cm of image_example.east] {\textbf{Vision}\\\textbf{Transformer}};
    \node[anchor=north west, xshift=-0.3cm, yshift=-0.1cm] at (image_encoder.north west) {\small\textcolor{white}{\faIcon[regular]{snowflake}
}};

    \node (structure) [draw, thick, rectangle, anchor=east] at (3, 2) {\footnotesize
        \textbf{Tree}\textbar{}\textcolor{gray!60}{Compact}\textbar{}\textcolor{gray!60}{Cups}\textbar{}\textcolor{gray!60}{Spider}
    };

    \node (tn) [draw, thick, right=0.5cm of text_encoder.east] {
\begin{tikzpicture}[
    scale=0.8,
    transform shape,
    tensor/.style={
      draw,
      circle,
      fill=TealGreen!50,
      minimum size=5mm,
      inner sep=0pt,
      thick,
    },
    edge/.style={
      thick,
    },
    every node/.style={font=\small}
  ]
    \node[tensor] (TA) at (0, 0) {};
    \node[tensor] (TL) at (1, 0) {};
    \node[tensor] (TB) at (2, 0) {};
    \node[tensor,fill=CoralRed] (TU1) at (1.5, -0.6) {};
    \node[tensor,fill=CoralRed] (TU2) at (0.75, -1.2) {};
    \node[above=2pt of TA] {Woman};
    \node[above=2pt of TL] {holds};
    \node[above=2pt of TB] {cat};
    \draw[edge] (TU2.south) -- ++(0,-0.4);

    \def\angleA{0}
    \def\angleB{180}
    \draw[edge] (TL.south) to[out=-90, in=\angleB] (TU1.\angleB);
    \draw[edge] (TB.south) to[out=-90, in=\angleA] (TU1.\angleA);
    \draw[edge] (TU1.south) to[out=-90, in=\angleA] (TU2.\angleA);
    \draw[edge] (TA.south) to[out=-90, in=\angleB] (TU2.\angleB);
\end{tikzpicture}
  };
        \def\boxsize{0.3cm}
        \definecolor{randcolor1}{RGB}{255, 99, 71}
        \definecolor{randcolor2}{RGB}{60, 179, 113}
        \definecolor{randcolor3}{RGB}{100, 149, 237}
        \definecolor{randcolor4}{RGB}{238, 130, 238}
        \definecolor{randcolor5}{RGB}{255, 215, 0}
        \definecolor{randcolor6}{RGB}{70, 130, 180}
        \definecolor{randcolor7}{RGB}{255, 140, 0}
        \foreach \i/\col in {-3/randcolor4, -2/randcolor1, -1/randcolor6, 0/randcolor2, +1/randcolor7, +2/randcolor3, +3/randcolor5} {
            \node (image_vector_\i) [rectangle, draw, fill=\col, minimum width=\boxsize, minimum height=\boxsize] at (13, -3cm + \i * \boxsize) {};
        }

        \node (tn_contract) [encoder, fill=MustardYellow, text=black, right=0.5cm of tn.east] {\textbf{TN}\\\textbf{Contractor}};

        \foreach \i/\col in {-3/randcolor2, -2/randcolor5, -1/randcolor1, 0/randcolor7, +1/randcolor4, +2/randcolor3, +3/randcolor6} {
            \node (text_vector_\i) [rectangle, draw, fill=\col, minimum width=\boxsize, minimum height=\boxsize] at (13, 0cm + \i * \boxsize) {};
        }

        \node[draw,thick,circle,fill=VibrantPurple,text=white] (cosine) at (14, -1.5) {cos};

        \draw[->, thick] (structure.east) -| (text_encoder.north);
        \draw[->, thick] (text_example.east) -- (text_encoder.west);
        \draw[->, thick] (image_example.east) -- (image_encoder.west);
        \draw[->, thick] (text_encoder.east) -- (tn.west);
        \draw[->, thick] (image_encoder.east) -- (image_vector_0.west);
        \draw[->, thick] (tn.east) -- (tn_contract.west);
        \draw[->, thick] (tn_contract.east) -- (text_vector_0.west);

        \draw[->, thick] (text_vector_0.east) -| (cosine.north);
        \draw[->, thick] (image_vector_0.east) -| (cosine.south);
        \draw[->, thick] (cosine.east) -- ++(0.5, 0);

        \coordinate (bbox-nw) at ([xshift=-0.5cm,yshift=0.5cm]text_encoder.north west);
        \coordinate (bbox-se) at ([xshift=0.8cm,yshift=-0.75cm]tn_contract.south east);
        \begin{scope}[on background layer]
        \draw[thick, dashed, fill=TealGreen!5] (bbox-nw) rectangle (bbox-se) node[pos=0.5, above, yshift=1.6cm] {\normalsize \textbf{Text Encoder}};
        \end{scope}
        \coordinate (bbox-nw-image) at ([xshift=-0.6cm,yshift=0.4cm]image_encoder.north west);
        \coordinate (bbox-se-image) at ([xshift=0.9cm,yshift=-0.65cm]image_encoder.south east);
        \begin{scope}[on background layer]
        \draw[thick, dashed, fill=CoralRed!5] (bbox-nw-image) rectangle (bbox-se-image) node[pos=0.5, below, yshift=-1.5cm] {\normalsize \textbf{Image Encoder}};
        \end{scope}

\end{tikzpicture}